\documentclass[sigconf]{acmart}
\usepackage{booktabs}
\usepackage{multirow}
\usepackage{subfigure}
\usepackage{balance}

\AtBeginDocument{%
  \providecommand\BibTeX{{%
    \normalfont B\kern-0.5em{\scshape i\kern-0.25em b}\kern-0.8em\TeX}}}

\copyrightyear{2022}
\acmYear{2022}
\setcopyright{acmcopyright}
\acmConference[MM '22]{Proceedings of the 30th ACM International Conference on Multimedia}{October 10--14, 2022}{Lisboa, Portugal}
\acmBooktitle{Proceedings of the 30th ACM International Conference on Multimedia (MM '22), October 10--14, 2022, Lisboa, Portugal}
\acmPrice{15.00}
\acmDOI{10.1145/3503161.3548166}
\acmISBN{978-1-4503-9203-7/22/10}

\begin{document}

\title{Look Before You Leap: Improving Text-based Person Retrieval by Learning A Consistent Cross-modal Common Manifold}

\author{Zijie Wang}
\affiliation{%
	\institution{Nanjing Tech University}
	\city{Nanjing}
	\country{China}}
\email{zijiewang9928@gmail.com}
\orcid{0000-0001-8739-7220}

\author{Aichun Zhu}
\authornote{Corresponding author.\vspace{-0.1cm}}
\affiliation{%
	\institution{Nanjing Tech University}
	\city{Nanjing}
	\country{China}}
\email{aichun.zhu@njtech.edu.cn}
\orcid{0000-0001-6972-5534}

\author{Jingyi Xue}
\affiliation{%
	\institution{Nanjing Tech University}
	\city{Nanjing}
	\country{China}}
\email{jyx981218@163.com}
\orcid{0000-0003-4889-6347}

\author{Xili Wan}
\affiliation{%
	\institution{Nanjing Tech University}
	\city{Nanjing}
	\country{China}}
\email{xiliwan@njtech.edu.cn}
\orcid{0000-0001-9160-8246}

\author{Chao Liu}
\affiliation{%
	\institution{Jinling Institute of Technology}
	\city{Nanjing}
	\country{China}}
\email{liuchao@jit.edu.cn}

\author{Tian Wang}
\affiliation{%
	\institution{Beihang University}
	\city{Beijing}
	\country{China}}
\email{wangtian@buaa.edu.cn}

\author{Yifeng Li}
\affiliation{%
	\institution{Nanjing Tech University}
	\city{Nanjing}
	\country{China}}
\email{lyffz4637@163.com}
\orcid{0000-0003-4798-3211}

\renewcommand{\shortauthors}{Zijie Wang et al.}

\begin{abstract}
  The core problem of text-based person retrieval is how to bridge the heterogeneous gap between multi-modal data. Many previous approaches contrive to learning a latent common manifold mapping paradigm following a \textbf{cross-modal distribution consensus prediction (CDCP)} manner. When mapping features from distribution of one certain modality into the common manifold, feature distribution of the opposite modality is completely invisible. That is to say, how to achieve a cross-modal distribution consensus so as to embed and align the multi-modal features in a constructed cross-modal common manifold all depends on the experience of the model itself, instead of the actual situation. With such methods, it is inevitable that the multi-modal data can not be well aligned in the common manifold, which finally leads to a sub-optimal retrieval performance. To overcome this \textbf{CDCP dilemma}, we propose a novel algorithm termed LBUL to learn a Consistent Cross-modal Common Manifold (C$^{3}$M) for text-based person retrieval. The core idea of our method, just as a Chinese saying goes, is to `\textit{san si er hou xing}', namely, to \textbf{Look Before yoU Leap (LBUL)}. The common manifold mapping mechanism of LBUL contains a looking step and a leaping step. Compared to CDCP-based methods, LBUL considers distribution characteristics of both the visual and textual modalities before embedding data from one certain modality into C$^{3}$M to achieve a more solid cross-modal distribution consensus, and hence achieve a superior retrieval accuracy. We evaluate our proposed method on two text-based person retrieval datasets CUHK-PEDES and RSTPReid. Experimental results demonstrate that the proposed LBUL outperforms previous methods and achieves the state-of-the-art performance.
\end{abstract}




\maketitle

\section{Introduction}

\begin{figure*}[h]
	\centering
	\includegraphics[width=\linewidth]{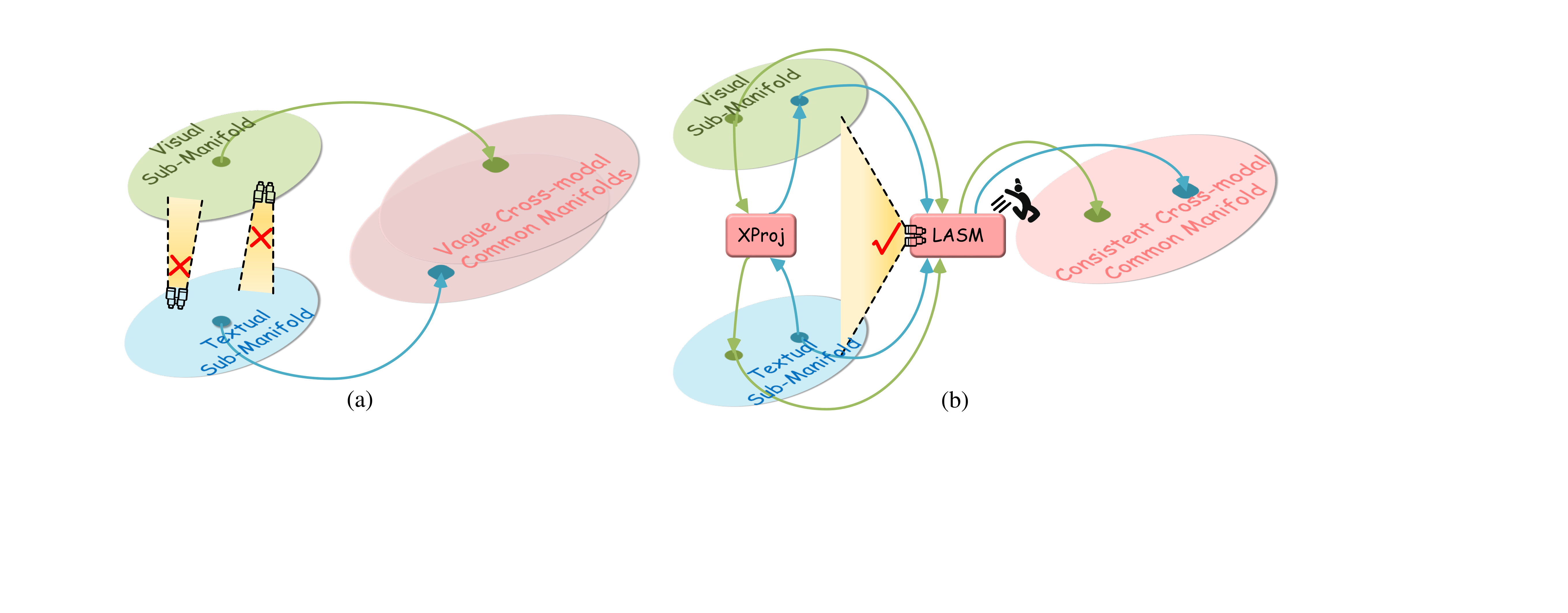}
	\caption{(a) For CDCP-based paradigms, when mapping features from distribution of one certain modality into the common manifold, feature distribution of the opposite modality is completely invisible. That is to say, how to achieve the cross-modal distribution consensus all depends on the experience of the model itself, instead of the actual situation. Consequently, there may exist multiple vague cross-modal common manifolds, which are adjacent to each other but not exactly identical. (b) For either a visual or textual sample, LBUL embeds it into C$^{3}$M after considering the distribution characteristics of both modalities to achieve a more solid cross-modal distribution consensus, instead of blindly predicting. }
	\label{fig:motivation}
\end{figure*}

Given a textual description query, text-based person retrieval aims to identify images of the corresponding pedestrian from a large-scale image database. Compared to the currently active research topic image-based person retrieval (aka. person re-identification)\cite{yi2014deepreid, IAM2019CVPR, SecondOrder2019} which utilizes image-based queries, text-based queries are much easier to access in the realistic application scenarios. Due to its effectiveness and applicability, text-based person retrieval \cite{Shuang2017Person, li2017identity, niu2020improving, jing2018pose, ARL, wang2020img, mm2019graphreid, wang2021amen, SUM} has drawn more and more attention. However, the study of this task is still in its infancy and there is still plenty of room for further research.

The core problem of text-based person retrieval is how to bridge the heterogeneous gap between multi-modal data. As different modalities are diverse and inconsistent in data form and distribution, it is not so easy to directly measure the cross-modal affinity. Many of the previous approaches contrive to learning a common manifold mapping paradigm, either implicitly with separate sub-models and extra constrains (e.g. attention mechanism), or explicitly with shared mapping blocks. These methods aim to transform heterogeneous multi-modal data into homogeneous feature representations, between which the cross-modal similarity can be calculated. However, most existing paradigms are proposed following a \textbf{cross-modal distribution consensus prediction (CDCP)} manner, which have their limitations. Specifically, as the multi-modal data are from specific distribution of each modality, the process of the common manifold mapping can be deemed as trying to achieve a distribution consensus between the visual and textual modalities, so as to embed and align the multi-modal features in a constructed cross-modal common manifold. Nevertheless, when mapping features from distribution of one certain modality into the common manifold, feature distribution of the opposite modality is completely invisible. That is to say,  how to achieve the cross-modal distribution consensus all depends on the experience of the model itself, instead of the actual situation. Consequently, as shown in Fig.~\ref{fig:motivation} (a), there may exist multiple vague cross-modal common manifolds, which are adjacent to each other but not exactly identical. With such methods, it is inevitable that the multi-modal data will not be perfected aligned and matched with each other in a proper common manifold, which finally leads to a sub-optimal retrieval performance. This situation can be called a \textbf{CDCP dilemma}.

To overcome this CDCP dilemma, we consider to come up with a more effective common manifold mapping paradigm. In this paper, we propose a novel algorithm to learn a Consistent Cross-modal Common Manifold (C$^{3}$M) for text-based person retrieval. As illustrated in Fig.~\ref{fig:motivation} (b), for either a visual or textual sample, our proposed method embeds it from one certain modality into C$^{3}$M after considering the distribution characteristics of both the visual and textual modalities to achieve a more solid cross-modal distribution consensus, instead of blindly predicting. The core idea of our method, just as a Chinese saying goes, is `san si er hou xing’, namely, to \textbf{Look Before yoU Leap}. So we name our proposed method \textbf{LBUL}, of which the common manifold mapping paradigm includes two steps, namely, a \textbf{looking step} and a \textbf{leaping step}. Compared with CDCP-based paradigms, LBUL is capable of achieving a more precise cross-modal distribution consensus. As a result, the multi-modal data can be embedded and aligned in a consistent common manifold with less information loss and higher accuracy, and hence achieve a superior retrieval performance. Specifically, with a proposed Uni-modal Sub-manifold Embedding Module (USEM), multi-granular features extracted from each modality are first distilled as a unified feature, which is embedded in a corresponding uni-modal sub-manifold (visual or textual). Then at the looking step of LBUL, in order to see distributions of both modalities before mapping data from one certain modality into C$^{3}$M, the uni-modal feature is projected into the opposite sub-manifold to give the distribution characteristics of the other modality by means of a Cross-modal Projection (XProj) module. In XProj, a Distribution Shifting ($\mathcal{DS}$) mechanism plays a key role in the statistical transformation of the feature according to the target modality, and thus enabling a proper feature projection. After the projection, for data from one certain modality, there exists two feature representations embedded in both the visual and textual modalities. Then at the leaping step, these two representations are processed together by a Leaping After Seeing Module (LASM), which conducts the common manifold mapping operation after seeing the distribution characteristics of both modalities. Through LASM, a consistent common representation can be obtained in C$^{3}$M for each sample, so that the cross-modal similarity can be properly measured. We evaluate our proposed method on two text-based person retrieval datasets including CUHK-PEDES \cite{Shuang2017Person} and RSTPReid \cite{dssl}. Experimental results demonstrate that LBUL outperforms previous methods and achieves the state-of-the-art performance.

The main contributions of this paper can be summarized as threefold: 
\begin{itemize}
	\item A novel LBUL method is proposed to learn a Consistent Cross-modal Common Manifold (C$^{3}$M) for text-based person retrieval, which embeds data from one certain modality into C$^{3}$M after considering the distribution characteristics of both the visual and textual modalities to achieve a more solid cross-modal distribution consensus, instead of blindly predicting.
	\item A two-step common manifold mapping mechanism which includes a looking step and a leaping step is proposed. By conducting the common manifold mapping operation after seeing the distribution characteristics of both modalities, LBUL is capable of learning consistent common representations with less information loss and higher retrieval accuracy.
	\item Extensive experimental analysis is carried out on CUHK-PEDES \cite{Shuang2017Person} and RSTPReid \cite{dssl} to evaluate the proposed LBUL method for text-based person retrieval. Experimental results demonstrate that LBUL significantly outperforms existing methods and achieves the state-of-the-art performance.
\end{itemize}

\section{Related Works}

\subsection{Person Re-identification}
Person re-identification has drawn increasing attention in both academical and industrial fields. This technology addresses the problem of matching pedestrian images across disjoint cameras. The key challenges lie in the large intra-class and small inter-class variation caused by different views, poses, illuminations, and occlusions. Existing methods can be grouped into handed-crafted descriptors, metric learning methods and deep learning methods. With the development of deep learning \cite{wang2021crossfoodTMM, sun2019supervisedhashingsigir, qian2021adaptiveTMM, zhao2021scalableKBS}, deep learning methods are in general playing a major role in current state-of-the-art works. Yi et al. \cite{yi2014deepreid} firstly proposed deep learning methods to match people with the same identification. To boost the ReID model training efficiency in multi-label classification, Wang et al. \cite{wang2020unsupervised} further proposed the memory-based multi-label classification loss (MMCL). MMCL works with memory-based non-parametric classifier and integrates multi-label classification and single-label classification in an unified framework. Jin et al. \cite{jin2020global} introduce a global distance-distributions separation (GDS) constraint over two distributions to encourage the clear separation of positive and negative samples from a global view.  Yuan et al. \cite{yuan2020deepgabor} propose a Gabor convolution module for deep neural networks based on Gabor function, which has a good texture representation ability and is effective when it is embedded in the low layers of a network. Taking advantage of the hinge function, they also design a new regularizer loss function to make the proposed Gabor Convolution module meaningful. A model that has joint weak saliency and attention aware is presented by Ning et al. \cite{ning2021jwsaa}, which can obtain more complete global features by weakening saliency features. In recent years, methods for unsupervised person re-identification have gradually emerged. Unsupervised person re-identification means that the target data set is unlabeled but the auxiliary source data set is not necessarily unlabeled \cite{fan2018unsupervised}. Existing unsupervised person ReID works can be concluded into three categories. The first category utilizes hand-craft features \cite{liao2015person}. But the features made by hand can not be robust and discriminative. To solve this problem, second category \cite{fu2019self} adopts clustering to estimate pseudo labels to train the CNN. However, these methods require good trained model. Recently, the third category is proposed, which improves unsupervised person ReID by using transfer learning. Some works \cite{wang2018transferable,lin2018multi} utilize transfer learning and minimize the attribute-level discrepancy by using extra attribute annotations.

\subsection{Text-based Person Retrieval}
Text-based person retrieval aims to search for the corresponding pedestrian image according to a given text query. This task is first introduced by Li et al. \cite{Shuang2017Person} and a GNA-RNN model is employed to handle the multi-modal data. Later, an efficient patch-word matching model \cite{Chen2018} is proposed to capture the local similarity between image and text. Jing et al. \cite{jing2018pose} utilize pose information as soft attention to localize the discriminative regions. Niu et al. \cite{niu2020improving} adopt a Multi-granularity Image-text Alignments (MIA) model exploit the combination of multiple granularities. Nikolaos et al. \cite{ARL} design a Text-Image Modality Adversarial Matching approach (TIMAM) to learn modality-invariant feature representation by means of adversarial and cross-modal matching objectives. Liu et al. \cite{mm2019graphreid} generate fine-grained structured representations from images and texts of pedestrians with an A-GANet model to exploit semantic scene graphs. CMAAM is introduced by Aggarwal et al. \cite{aggarwal2020text} which learns an attribute-driven space along with a class-information driven space by introducing extra attribute annotation and prediction. Zheng et al. \cite{zheng2020hierarchical} propose a Gumbel attention module to alleviate the matching redundancy problem and a hierarchical adaptive matching model is employed to learn subtle feature representations from three different granularities. Zhu et al. \cite{dssl} proposed a Deep Surroundings-person Separation Learning (DSSL) model to effectively extract and match person information. Besides, they construct a Real Scenarios Text-based Person Re-identification (RSTPReid) dataset based on MSMT17 \cite{MSMT17} to benefit future research on text-based person retrieval. Most of the above mentioned approaches are proposed following a cross-modal distribution consensus prediction (CDCP) manner. By means of the LBUL mechanism, a more solid cross-modal distribution consensus can be achieved and hence a more consistent cross-modal common manifold can be constructed.

\begin{figure*}[h]
	\centering
	\includegraphics[width=\linewidth]{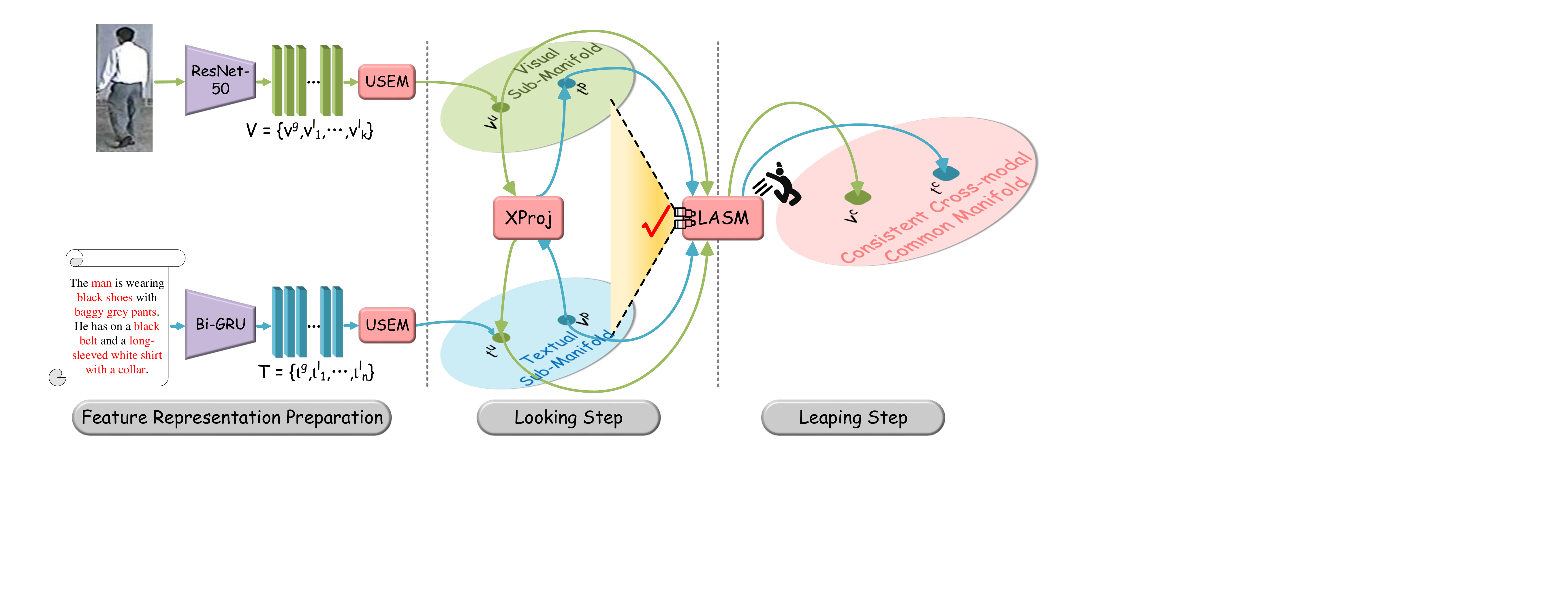}
	\caption{The overall framework of the proposed LBUL model.}
	\label{fig:model}
\end{figure*}

\section{Methodology}

\subsection{Problem Formulation}
The goal of the proposed framework (shown in Fig.~\ref{fig:model}) is to measure the similarity between multi-modal data, namely, a given textual description query and a gallery person image. Formally, let $D = \{p_{i}, q_{i}\}_{i=1}^{N}$ denotes a dataset consists of $N$ image-text pairs. Each pair contains a pedestrian image $p_{i}$ captured by one certain surveillance camera and its corresponding textual description query $q_{i}$. The IDs of the $Q$ pedestrians in the dataset are denoted as $Y = \{y_{i}\}_{i=1}^{Q}$. Given a textual description, the aim is to identify images of the most relevant pedestrian from a large scale person image gallery.

\subsection{Feature Representation Preparation}

\subsubsection{\textbf{Representation Extraction}}

\paragraph{\textbf{Visual Representation Extraction}}
To extract multi-granular visual representations from a given image $I$, a pretrained ResNet-50 \cite{ResNet} backbone is utilized. To obtain the global representation $v^{g} \in \mathbb{R}^{p}$, the feature map before the last pooling layer of ResNet-50 is down-scaled to $1 \times 1 \times 2048$ with an average pooling layer and converted into a $2048$-dim vector. Then it is passed through a group normalization (GN) layer followed by a fully-connected (FC) layer and transformed to $p$-dim. In the local branch, the same feature map is first horizontally $k$-partitioned by pooling it to $k \times 1 \times 2048$, and then the local strips are separately passed through a GN and two FCs with an ELU layer between them to form $k$ $p$-dim vectors $V^{l} = \{v^{l}_{1}, v^{l}_{2}, \cdots, v^{l}_{k}\}$, which are finally concatenated with each other along with $v^{g}$ to obtained the visual representation matrix $V = \{v^{g}, v^{l}_{1}, v^{l}_{2}, \cdots, v^{l}_{k}\} \in \mathbb{R}^{p \times (k + 1)}$.

\paragraph{\textbf{Textual Representation Extraction}}
For textual representation extraction, a whole sentence along with $n$ phrases extracted from it is taken as textual materials, which is processed by a bi-directional Gated Recurrent Unit (bi-GRU). The last hidden states of the forward and backward GRUs are concatenated to give global/local $2p$-dim feature vectors. And then the $2p$-dim vector got from the whole sentence is passed through a GN followed by an FC to form the global textual representation $t^{g} \in \mathbb{R}^{p}$. With each certain input phrase, the corresponding output $p$-dim vector is handled consecutively by a GN and two FCs with an ELU layer between them. The obtained local vectors $T^{l} = \{t^{l}_{1}, t^{l}_{2}, \cdots, t^{l}_{n}\}$ are then concatenated with each other along with $t^{g}$ to form the final textual representation matrix $T = \{t^{g}, t^{l}_{1}, t^{l}_{2}, \cdots, t^{l}_{n}\} \in \mathbb{R}^{p \times (n + 1)}$.

\subsubsection{\textbf{Uni-modal Sub-manifold Learning}}
After the raw visual and textual representation matrices $V$ and $T$ are obtained, we adopt a Uni-modal Sub-manifold Embedding Module (USEM) based on the self-attention mechanism \cite{vaswani2017attentionBERT} to distill both global and fine-grained local discriminative information into a unified feature, which can be formulated as
\begin{equation}
	v^{u} = USEM(V) = \sum_{i = 1}^{k} \frac{exp(v^g v_i^l)}{\sum_{j=1}^k exp(v^g v_j^l)}v^{l}_{i} + v^g,
\end{equation}
\begin{equation}
	t^{u} = USEM(T) = \sum_{i = 1}^{n} \frac{exp(t^g t_i^l)}{\sum_{j=1}^n exp(t^g t_j^l)}t^{l}_{i} + t^g,
\end{equation}
where $v^{u}$ or $t^{u}$ is the distilled unified visual/textual feature and is embedded into the corresponding visual/textual sub-manifold, respectively.

\subsection{Look Before You Leap}
Now that the uni-modal representations are obtained and each uni-modal sub-manifold which reveals the latent distribution characteristics of the corresponding modality is constructed, a two-step common manifold mapping mechanism including a looking step and a leaping step is proposed, which aims to embed the multi-modal data into the consistent cross-modal common manifold C$^{3}$M with less information loss and higher accuracy.

\subsubsection{\textbf{Looking Step: Cross-modal Projection Module}}
At the looking step, the prime target is to see distributions of both modalities before mapping data from one certain modality into C$^{3}$M. To achieve this goal, a Cross-modal Projection (XProj) module is proposed to project one certain uni-modal representation into the opposite sub-manifold, which can be generally formulated as
\begin{equation}
	s^{p} = XProj(s^{u}, r^{u}),
\end{equation}
where $s^{u}$ or $r^{u}$ denotes the input source or target modal representation, respectively.  $s^{p}$ is the projected representation, which is embedded into the target sub-manifold.

Specifically, a Distribution Shifting ($\mathcal{DS}$) mechanism is first adopted to conduct the statistical transformation of the source modal representation according to the target modality:
\begin{equation}
	s^{ds} = \mathcal{DS}(s^{u}, r^{u}) = \sigma (r^{u}) (\frac{s^{u} - \mu (s^{u})}{\sigma (s^{u})}) + \mu (r^{u}),
\end{equation}
where $s^{ds}$ is the shifted feature. $\mu(\cdot)$ and $\sigma(\cdot)$ denote the calculation of mean and variance, respectively. Then a multi-layer perceptron ($\mathcal{MLP}$) with a tanh activation layer is employed to embed $s^{ds}$ into the target sub-manifold:
\begin{equation}
	s^{p} = \mathcal{MLP}(s^{ds}).
\end{equation}
With XProj, representation lying in one certain uni-modal sub-manifold can be properly projected into the opposite sub-manifold:
\begin{equation}
	v^{p} = XProj(v^{u}, t^{u}), \ t^{p} = XProj(t^{u}, v^{u}).
\end{equation}

\subsubsection{\textbf{Leaping Step: Leaping After Seeing Module}}

After the looking step, for data from one certain modality, there exists two feature representations embedded in both the visual and textual modalities, which give the distribution characteristics of both modalities. Then at the leaping step, the two representations can be utilized by LBUL with a proposed Leaping After Seeing Module (LASM) to conduct the common manifold mapping operation after seeing distributions of both modalities. The mechanism of LASM can be formulated as
\begin{equation}
	x^{c} = LASM(x^{u}, x^{p}),
\end{equation}
where $x^{u}$ and $x^{p}$ are visual/textual representations before and after processed by XProj, respectively, while $x^{c}$ is the visual/textual consistent common manifold representation embedded in C$^{3}$M.

To be specific, the two input representations are first utilized to estimate a fusion gate $g \in \mathbb{R}^{p}$:
\begin{equation}
	\label{eq:oplus}
	g = \sigma(W_{2}ELU(W_{1}(x^{u} \oplus x^{p}))),
\end{equation}
where $\oplus$ is the feature concatenation operation and can be implemented as several other methods (e.g. addition or concatenation), which will be further discussed along with some substitution variants of LASM in Sec.~\ref{sec:exp_lasm}. Here, $W_{1} \in \mathbb{R}^{2p \times 2p}$ and $W_{2} \in \mathbb{R}^{2p \times p}$ denote linear transformations without bias while $\sigma(\cdot)$ stands for the sigmoid activation function. Then $x^{c}$ is obtained through the weighted summation of $x^{u}$ and $x^{p}$ according to $g$:
\begin{equation}
	\label{eq:lasm}
	x^{c} = g x^{u} + (1 - g) x^{p}.
\end{equation}
To sum up, the corresponding visual and textual consistent cross-modal common manifold representations $v^{c}$ and $t^{c}$ can be obtained as
\begin{equation}
	v^{c} = LASM(v^{u}, v^{p}), \ t^{c} = LASM(t^{u}, t^{p}).
\end{equation}

\subsection{Similarity for Inference}
For test and inference, several kinds of similarity scores are calculated. First, the similarity $sim^{c}$ between a pair of visual/textual representations embedded in C$^{3}$M is calculated:
\begin{equation}
	sim^{c} = cos(v^{c}, t^{c}),
\end{equation}
where $cos(\cdot, \cdot)$ denotes the cosine similarity between two feature vectors. To further improve the retrieval performance, the global and fine-grained similarities are also utilized as an auxiliary. The global similarity $sim^{g}$ between a pair of global multi-modal representations is computed as
\begin{equation}
	sim^{g} = cos(v^{g}, t^{g}).
\end{equation}
To obtain the fine-grained similarity $sim^{f}$, a cross-modal attention ($\mathcal{CA}$) mechanism is adopted:
\begin{equation}
	\alpha^{X}_{i} = \frac{exp(cos(x^{l}_{i}, y^{g}))}{\sum_{j} exp(cos(X^{l}_{j}, y^{g}))},
\end{equation}
\begin{equation}
	x^{f} = \mathcal{CA}(y^{g}, X^{l}) = \sum\limits_{\alpha^{X}_{i} > \gamma} \alpha^{X}_{i} x^{l}_{i},
\end{equation}
where $(X, Y)$ can be $(V, T)$ or $(T, V)$ while $(x, y)$ can be $(v, t)$ or $(t, v)$. $\gamma$ is a threshold value. Thus, $sim^{f}$ can be calculated by
\begin{equation}
	v^{f} = \mathcal{CA}(t^{g}, V^{l}), \ t^{f} = \mathcal{CA}(v^{g}, T^{l}),
\end{equation} 
\begin{equation}
	sim^{f} = \frac{cos(v^{g}, t^{f}) + cos(v^{f}, t^{g})}{2}.
\end{equation}
Eventually, we can get the overall similarity $sim$ as
\begin{equation}
	sim = sim^{c} + \lambda_{1} sim^{g} + \lambda_{2} sim^{f}.
\end{equation}

\subsection{Optimization}
\label{sec:opti}
To optimize LBUL, the ranking loss is employed to constrain the matched pairs to be closer than the mismatched ones in a mini-batch with a margin $\beta$:
\begin{multline}
	L_{rk}(x_1, x_2) = \sum_{\widehat{x_2}} max\{\beta - cos(x_1, x_2) + cos(x_1, \widehat{x_2}), 0\}\\
	+ \sum_{\widehat{x_1}} max\{\beta - cos(x_1, x_2) + cos(\widehat{x_1}, x_2), 0\},
\end{multline}
where $(x_1, \widehat{x_2})$ or $(\widehat{x_1}, x_2)$ denotes a mismatched pair while $(x_1, x_2)$ is a matched pair. Besides, the identification (ID) loss is also adopted:
\begin{equation}
	L_{id}(x) =  -log(softmax(W_{id} x),
\end{equation}
where $W_{id} \in \mathbb{R}^{Q \times p}$ is a shared FC layer without bias while $Q$ is the number of different pedestrians.

The overall optimization process of LBUL includes two stages. Before conducting the common manifold mapping operation, it is necessary to ensure each learned sub-manifold is solid. Therefore, in the first stage, the ranking loss and ID loss are utilized to optimize the extracted global and fine-grained local features:
\begin{equation}
	\mathcal{L}^{g} = L_{id}(v^{g}) + L_{id}(t^{g}) + L_{rk}(v^{g}, t^{g}),
\end{equation}
\begin{equation}
	\mathcal{L}^{f} = L_{id}(v^{f}) + L_{id}(t^{f}) + L_{rk}(v^{g}, t^{f}) + L_{rk}(v^{f}, t^{g}),
\end{equation}
\begin{equation}
	\mathcal{L}^{Stage1} = \mathcal{L}^{g} + \lambda_{3}\mathcal{L}^{f}.
\end{equation}
In the second stage, first the ranking loss between the projected feature and the unified feature in the target modality is computed to ensure a reliable projection. Besides, the ID loss is also employed:
\begin{multline}
	\mathcal{L}^{p} = L_{id}(v^{u}) + L_{id}(t^{u}) + L_{id}(v^{p}) + L_{id}(t^{p}) \\ + L_{rk}(v^{p}, t^{u}) + L_{rk}(v^{u}, t^{p}),
\end{multline}
And then the loss between a pair of common manifold features is calculated as
\begin{equation}
	\mathcal{L}^{c} = L_{id}(v^{c}) + L_{id}(t^{c}) + L_{rk}(v^{c}, t^{c}).
\end{equation}
The entire loss for the second stage is
\begin{equation}
	\mathcal{L}^{Stage2} = \mathcal{L}^{Stage1} + \lambda_{4}\mathcal{L}^{p} + \lambda_{5}\mathcal{L}^{c}.
\end{equation}

\section{Experiments}

\subsection{Experimental Setup}

\subsubsection{\textbf{Dataset and Metrics}}
Our approach is evaluated on two challenging Text-based Person Retrieval datasets including CUHK-PEDES \cite{Shuang2017Person} and RSTPReid \cite{dssl}.

(1) \textbf{CUHK-PEDES}: Following the official data split approach \cite{Shuang2017Person}, the training set of CUHK-PEDES contains 34054 images, 11003 persons and 68126 textual descriptions. The validation set contains 3078 images, 1000 persons and 6158 textual descriptions while the test set has 3074 images, 1000 persons and 6156 descriptions. Every image generally has two descriptions, and each sentence is commonly no shorter than 23 words. After dropping words that appear less than twice, the word number is 4984.

(2) \textbf{RSTPReid}: The RSTPReid dataset \cite{dssl} is constructed based on MSMT17 \cite{MSMT17}, which contains 20505 images of 4,101 persons from 15 cameras. Each person has 5 corresponding images taken by different cameras and each image is annotated with 2 textual descriptions. For data division, 3701, 200 and 200 identities are utilized for training, validation and test, respectively. Each sentence is no shorter than 23 words.

\paragraph{\textbf{Evaluation Metrics}} The performance is evaluated by the rank-k accuracy. All images in the test set are ranked by their similarities with a given query natural language sentence. If any image of the corresponding person is contained in the top-k images, we call this a successful search. We report the rank-1, rank-5, and rank-10 accuracies for all experiments.

\subsubsection{\textbf{Implementation Details}}
In our experiments, we set the representation dimensionality $p = 2048$. The dimensionality of embedded word vectors is set to 500. The pretrained ResNet-50 \cite{ResNet} is utilized as the visual CNN backbone and a pretrained BERT language model \cite{vaswani2017BERT} is used to better handle the textual input. The total number of noun phrases obtained from each sentence is kept flexible while $k$ is set to 6. For both the CUHK-PEDES and RSTPReid dataset, the input images are resized to $384 \times 128 \times 3$. The random horizontal flipping strategy is employed for data augmentation. The threshold value $\gamma$ in $\mathcal{CA}$ can be $\frac{1}{k}$ or $\frac{1}{n}$ for visual or textual data, respectively. The margin $\beta$ of ranking losses is set to 0.2 while the $\lambda$'s are empirically set to 1 in this paper. An Adam optimizer \cite{AdamOptimizer} is adopted to train the model with a batch size of 64 for 100 epochs.

\subsection{Ablation Analysis}
\label{sec:abla}

\begin{table}[!ht]
	\caption{Performance comparisons of common manifold mapping paradigms on CUHK-PEDES and RSTPReid.}
	\label{tab:baseline}
	\centering
	\begin{tabular}{c|l|ccc}
		\toprule
		~& Method & Rank-1 & Rank-5 & Rank-10 \\
		\midrule
		\multirow{6}{*}{\rotatebox{90}{CUHK-PEDES}} & CDCP-Sep (glo) & 56.19 & 76.43 & 83.63 \\
		~& CDCP-Sha (glo)  & 56.78 & 77.19 & 84.07 \\
		~& LBUL (glo) & \textbf{57.20} & \textbf{77.78} & \textbf{84.23} \\
		\cline{2-5}
		~& CDCP-Sep (USEM) & 59.62 & 79.39 & 85.83 \\
		~& CDCP-Sha (USEM) & 59.88 & 79.56 & 85.91 \\
		~& LBUL (USEM)& \textbf{61.95} & \textbf{81.16} & \textbf{87.19} \\
		\midrule
		\multirow{6}{*}{\rotatebox{90}{RSTPReid}} & CDCP-Sep (glo) & 37.05 & 61.75 & 71.10 \\
		~& CDCP-Sha (glo) & 37.85 & 62.05 & 71.50 \\
		~& LBUL (glo) & \textbf{38.65} & \textbf{64.70} & \textbf{73.20} \\
		\cline{2-5}
		~& CDCP-Sep (USEM) & 40.70 & 65.55 & 75.10 \\
		~& CDCP-Sha (USEM) & 41.40 & 65.85 & 75.40 \\
		~& LBUL (USEM) & \textbf{43.35} & \textbf{66.85} & \textbf{76.50} \\
		\bottomrule
	\end{tabular}
\end{table}

\subsubsection{\textbf{Comparison with CDCP-based Paradigms}}

The core idea of this paper is the `Look Before You Leap (LBUL)' mechanism. To demonstrate its effectiveness, a series of experiments on CUHK-PEDES and RSTPReid are carried out to compare LBUL with CDCP-based paradigms:

(1) \textbf{CDCP-Sep (glo)}: directly using the global features $v^g$ and $t^g$ extracted by modality-specific sub-models to calculate $s^{g}$ for matching;

(2) \textbf{CDCP-Sha (glo)}: calculating $s^{g}$ after processing $v^g$ and $t^g$ with a shared mapping block;

(3) \textbf{LBUL (glo)}: processing global features with LBUL mechanism (XProj + LASM) to obtain $s^{g}$;

(4) \textbf{CDCP-Sep (USEM)}: calculating a similarity $s^u$ with output of USEM $v^u$ and $t^u$, and then matching multi-modal samples with $s^g$, $s^f$ and $s^u$;

(5) \textbf{CDCP-Sha (USEM)}: calculating $s^c$ after processing $v^u$ and $t^u$ with a shared mapping block and then matching multi-modal samples with $s^g$, $s^f$ and $s^c$;

(6) \textbf{LBUL (USEM)}: just equivalent to the complete LBUL method, which conducts the LBUL-based operation on $v^u$ and $t^u$ to obtain $s^c$ and matches multi-modal samples with $s^g$, $s^f$ and $s^c$.

The experimental results are reported in Tab.~\ref{tab:baseline}. As can be observed from the table, for CDCP-based methods, approaches using a shared common manifold mapping block outperform ones without to some extent. Furthermore, with the idea of LBUL for common manifold mapping, an obvious performance gain can be achieved. For instance, for methods using $v^u$ and $t^u$, the LBUL-based method outperforms the CDCP-based method with a shared mapping block by $2.07\%$, $1.60\%$, $1.28\%$ and $1.95\%$, $1.00\%$, $1.10\%$ on CUHK-PEDES and RSTPReid, respectively. The experimental results demonstrate that by conducting the common manifold mapping operation after seeing distribution characteristics of both modalities, LBUL is more capable of learning consistent common representations with less information loss and higher retrieval accuracy.

\begin{table*}[!ht]
	\caption{Ablation study on proposed components of LBUL on CUHK-PEDES and RSTPReid.}
	\label{tab:alba0}
	\centering
	\begin{tabular}{ ccc | c | c | c | c | ccc | ccc}
		\toprule
		\multicolumn{7}{c}{Component}& \multicolumn{3}{|c|}{CUHK-PEDES}& \multicolumn{3}{|c}{RSTPReid} \\
		\midrule
		$s^{g}$ & $s^{f}$ & $s^{c}$ & 2ST & USEM & LASM & BERT & Rank-1 & Rank-5 & Rank-10 & Rank-1 & Rank-5 & Rank-10 \\
		\midrule
		$\checkmark$ & $\checkmark$ & $\checkmark$ & $\times$ & $\checkmark$ & $\checkmark$ & $\times$ & 53.50 & 75.05 & 82.68 & 34.85 & 60.15 & 70.95 \\
		\midrule
		$\checkmark$ & $\times$ & $\times$ & - & - & - & $\times$ & 54.81 & 75.65 & 83.43 & 37.00 & 61.75 & 71.10 \\
		$\checkmark$ & $\checkmark$ & $\times$ & - & - & - & $\times$ & 57.37 & 77.83 & 85.00 & 40.15 & 63.95 & 74.10 \\
		\midrule
		$\checkmark$ & $\checkmark$ & $\checkmark$ & $\checkmark$& glo & $\checkmark$ & $\times$ & 60.81 & 80.79 & 86.72 & 41.60 & 65.55 & 75.30 \\
		$\checkmark$ & $\checkmark$ & $\checkmark$ & $\checkmark$& avg & $\checkmark$ & $\times$ & 60.37 & 80.17 & 86.62 & 42.40 & 66.15 & 76.10 \\
		$\checkmark$ & $\checkmark$ & $\checkmark$ & $\checkmark$& avgloc + glo & $\checkmark$ & $\times$ & 61.04 & 80.95 & 86.85 & 42.15 & 66.00 & 76.15 \\
		\midrule
		$\checkmark$ & $\checkmark$ & $\checkmark$ & $\checkmark$ & $\checkmark$ & w/o $\mathcal{DS}$ & $\times$ & 60.24 & 80.12 & 86.32 & 41.25 & 65.15 & 75.55 \\
		$\checkmark$ & $\checkmark$ & $\checkmark$ & $\checkmark$ & $\checkmark$ & add & $\times$ & 60.57 & 80.56 & 86.56 & 40.45 & 64.35 & 73.70\\
		$\checkmark$ & $\checkmark$ & $\checkmark$ & $\checkmark$ & $\checkmark$ & add + $\mathcal{MLP}$ & $\times$ & 60.32 & 80.05 & 86.53 & 41.15 & 65.25 & 75.35\\
		$\checkmark$ & $\checkmark$ & $\checkmark$ & $\checkmark$ & $\checkmark$ & concat & $\times$ & 60.31 & 80.41 & 86.93 & 42.25 & 66.30 & 76.15 \\
		$\checkmark$ & $\checkmark$ & $\checkmark$ & $\checkmark$ & $\checkmark$ & concat + $\mathcal{MLP}$ & $\times$ & 60.39 & 80.36 & 86.19 & 42.05 & 65.70 & 75.95 \\
		$\checkmark$ & $\checkmark$ & $\checkmark$ & $\checkmark$ & $\checkmark$ & scalar gate & $\times$ & 61.27 & 80.70 & 86.89 & 42.65 & 66.55 & 76.30 \\
		$\checkmark$ & $\checkmark$ & $\checkmark$ & $\checkmark$ & $\checkmark$ & $\oplus$=add & $\times$ & 61.74 & 80.93 & 86.97 & 43.20 & 66.60 & \underline{76.55}\\
		\midrule
		$\checkmark$ & $\checkmark$ & $\checkmark$ & $\checkmark$ & $\checkmark$ & $\checkmark$ & $\times$ & \underline{61.95} & \underline{81.16} & \underline{87.19} & \underline{43.35} & \underline{66.85} & 76.50 \\
		$\checkmark$ & $\checkmark$ & $\checkmark$ & $\checkmark$ & $\checkmark$ & $\checkmark$ & $\checkmark$ & \textbf{64.04} & \textbf{82.66} & \textbf{87.22} & \textbf{45.55} & \textbf{68.20} & \textbf{77.85} \\
		\bottomrule
	\end{tabular}
\end{table*}

\subsubsection{\textbf{Impact of Uni-modal Sub-manifold Embedding Module (USEM)}}

We enumerate some variants for the Uni-modal Sub-manifold Embedding Module (USEM) and compare them with our proposed USEM:

(1) \textbf{Glo}: $v^u, \ t^u = v^g, \ t^g$;

(2) \textbf{Avg}: $v^u, \ t^u = avg(V), \ avg(T)$;

(3) \textbf{AvgLoc + Glo}: $v^u, \ t^u = v^g + avg(V^l), \ t^g + avg(T^l)$, where $avg(\cdot)$ denotes the feature averaging operation.

We can observe from Tab.~\ref{tab:alba0} that the performance of the model with USEM is obviously better than the other substitutions, which indicate the effectiveness of USEM.

\begin{table}[!ht]
	\caption{Comparison with SOTA on CUHK-PEDES.}
	\label{tab:sota}
	\centering
	\begin{tabular}{l|ccc}
		\toprule
		Method & Rank-1 & Rank-5 & Rank-10 \\
		\midrule
		CNN-RNN \cite{reed2016learning}& 8.07  & -  & 32.47 \\
		Neural Talk \cite{vinyals2015show} & 13.66  & -  & 41.72 \\
		GNA-RNN \cite{Shuang2017Person} & 19.05  & -  & 53.64 \\
		IATV \cite{li2017identity} & 25.94  & -  & 60.48 \\
		PWM-ATH \cite{Chen2018} & 27.14  & 49.45  & 61.02 \\
		Dual Path \cite{zheng2020dual} & 44.40  & 66.26  & 75.07 \\
		GLA \cite{chen2018improving} & 43.58  & 66.93  & 76.26 \\
		CMPM-CMPC  \cite{zhang2018CMPM-CMPC} & 49.37 & 71.69 & 79.27 \\
		MIA \cite{niu2020improving} & 53.10 & 75.00 & 82.90 \\
		A-GANet \cite{mm2019graphreid} & 53.14 & 74.03 & 81.95 \\
		PMA \cite{jing2018pose} & 54.12 & 75.45 & 82.97 \\
		TIMAM \cite{ARL} & 54.51 & 77.56 & 84.78 \\
		CMAAM \cite{aggarwal2020text} & 56.68 &	77.18 &	84.86 \\
		AMEN \cite{wang2021amen} & 57.16 & 78.64& 86.22 \\
		HGAN \cite{zheng2020hierarchical} & 59.00 & 79.49 & 86.62 \\
		DSSL \cite{dssl} & 59.98 & 80.41 & \underline{87.56} \\
		\midrule
		\textbf{LBUL (Ours)} & \underline{61.95} & \underline{81.16} & 87.19 \\
		\textbf{LBUL + BERT (Ours)} & \textbf{64.04} & \textbf{82.66} & \textbf{87.22} \\
		\bottomrule
	\end{tabular}
\end{table}

\begin{table}[!ht]
	\caption{Comparison with SOTA on RSTPReid.}
	\label{tab:sota_retpreid}
	\centering
	\begin{tabular}{l|ccc}
		\toprule
		Method & Rank-1 & Rank-5 & Rank-10 \\
		\midrule
		IMG-Net \cite{wang2020img} & 37.60 & 61.15 & 73.55 \\
		AMEN \cite{wang2021amen} & 38.45 & 62.40 & 73.80 \\
		DSSL \cite{dssl} & 39.05 & 62.60 & 73.95 \\
		SSAN \cite{ding2021ssan} & 43.50 & 67.80 & 77.15 \\
		\midrule
		\textbf{LBUL (ours)} & \underline{43.35} & \underline{66.85} & \underline{76.50} \\
		\textbf{LBUL + BERT (Ours)} & \textbf{45.55} & \textbf{68.20} & \textbf{77.85} \\
		\bottomrule
	\end{tabular}
\end{table}

\subsubsection{\textbf{Impact of Leaping After Seeing Module (LASM)}}
\label{sec:exp_lasm}
How to properly take advantage of the two feature representations $x^u$ and $x^p$ in the Leaping After Seeing Module (LASM) is crucial to the performance of LBUL. Therefore, we conduct experiments with several substitution variants of LASM on both the CUHK-PEDES and RSTPReid dataset to see the effectiveness of our proposed method:

(1) \textbf{Add}: $x^c = x^u + x^p$;

(2) \textbf{Add + $\mathcal{MLP}$}: $x^c = tanh(W(x^u + x^p) + b)$;

(3) \textbf{Concat}: $x^c = \left[\begin{matrix} x^u \\ x^p \end{matrix}\right]$;

(4) \textbf{Concat + $\mathcal{MLP}$}: $x^c = tanh(W\left[\begin{matrix} x^u \\ x^p \end{matrix}\right] + b)$;

(5) \textbf{Scalar Gate}: $x^c = a x^u + (1- a) x^p$ where $a$ is a real-valued number parameterized by $x^u$ and $x^p$;

(6) \textbf{$\oplus$=add}: substitute the concatenation operation in Eq.~\ref{eq:oplus} for addition.

As can be seen from Tab.~\ref{tab:alba0}, the feature aggregation paradigm proposed in LASM (i.e. Eq.~\ref{eq:oplus} and Eq.~\ref{eq:lasm}) achieves substantially better performance than all the basic variants, while implementing $\oplus$ in Eq.~\ref{eq:oplus} as addition or concatenation give similar retrieval accuracies, with the concatenation method slightly better.

\begin{figure*}[!ht]
	\centering
	\subfigure[Rank-1 Accuracy]{
		\begin{minipage}[t]{0.25\linewidth}
			\centering
			\includegraphics[width=\linewidth]{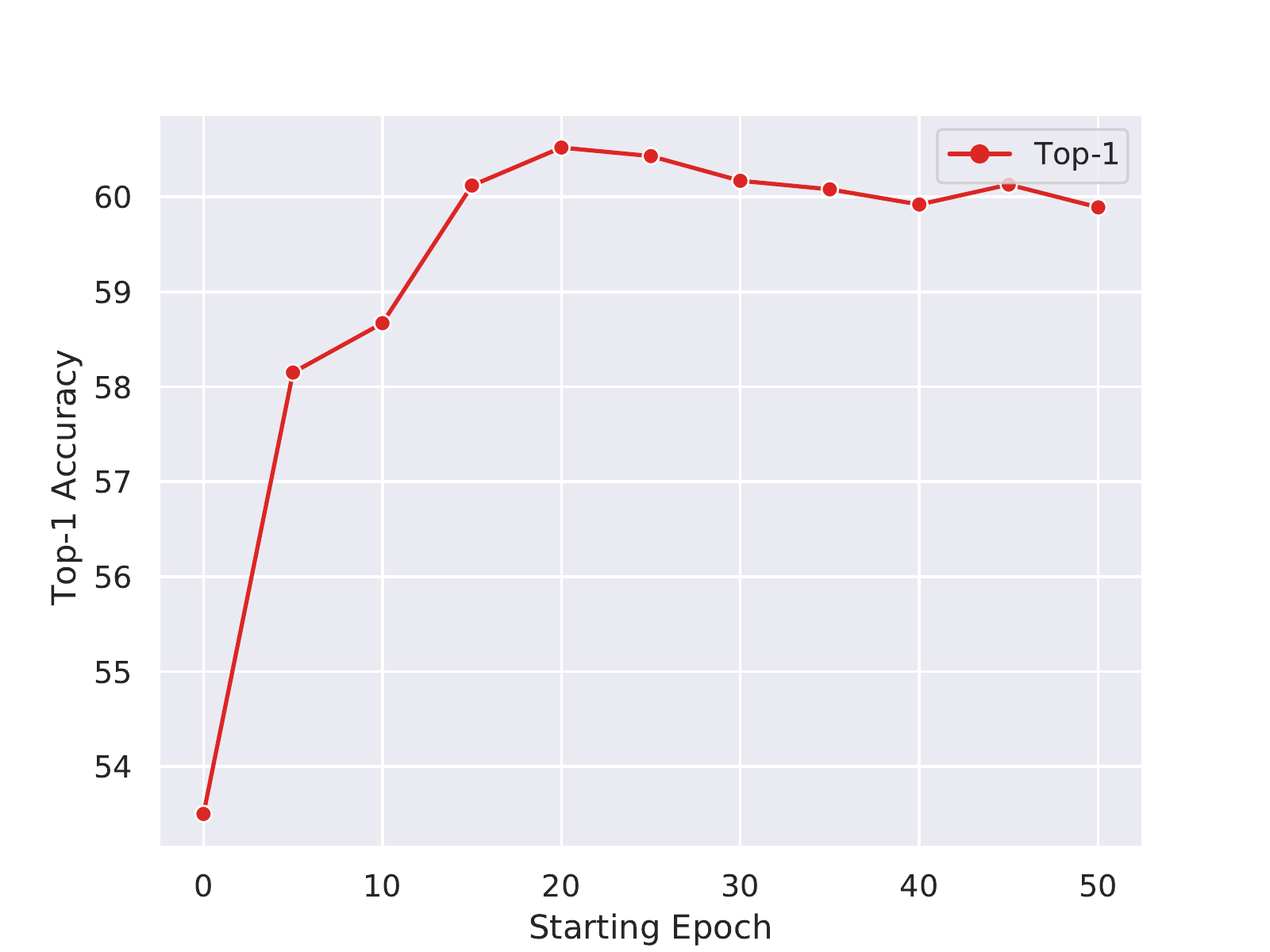}
		\end{minipage}%
	}%
	\subfigure[Rank-5 Accuracy]{
		\begin{minipage}[t]{0.25\linewidth}
			\centering
			\includegraphics[width=\linewidth]{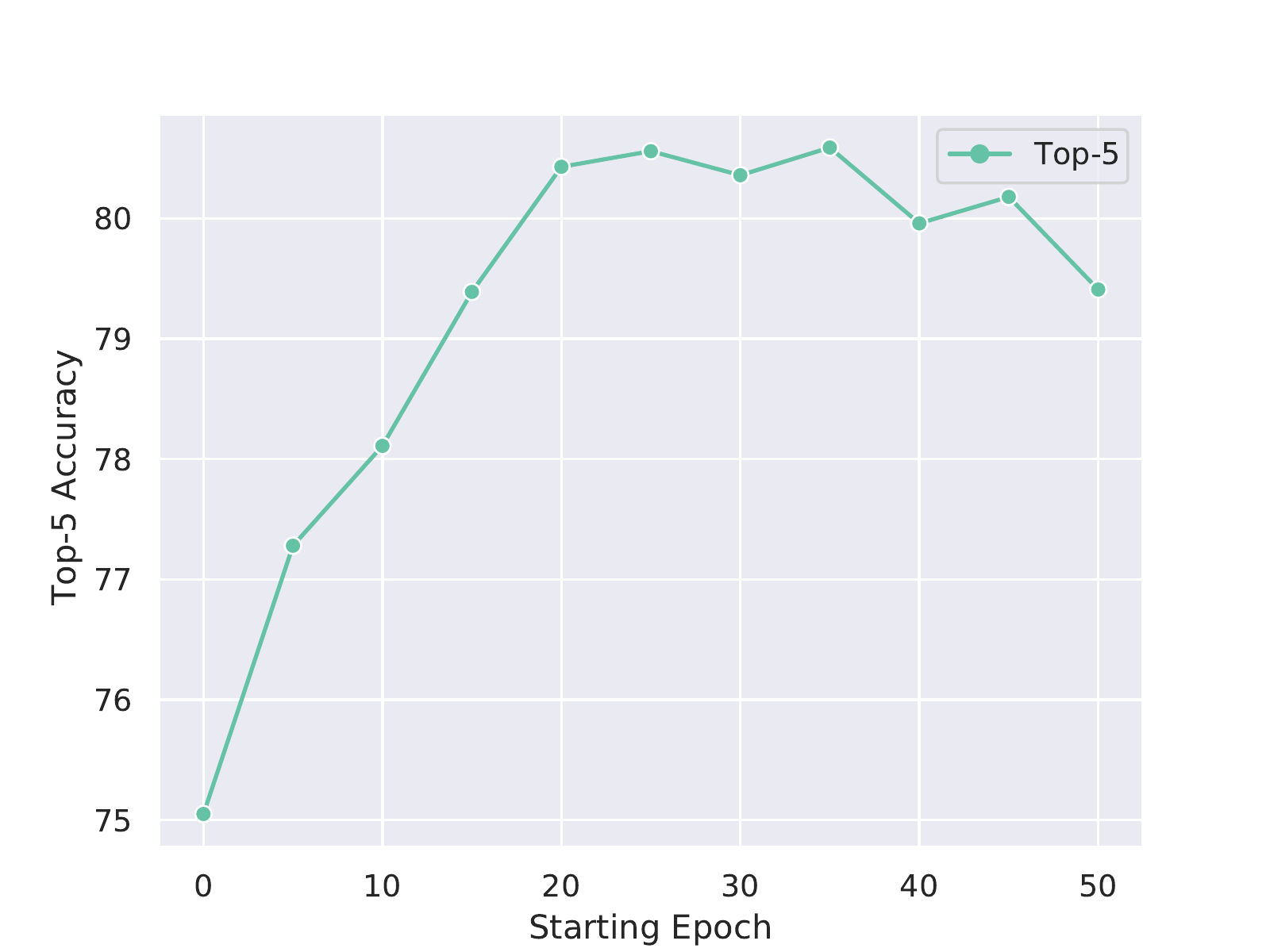}
		\end{minipage}%
	}%
	\subfigure[Rank-10 Accuracy]{
		\begin{minipage}[t]{0.25\linewidth}
			\centering
			\includegraphics[width=\linewidth]{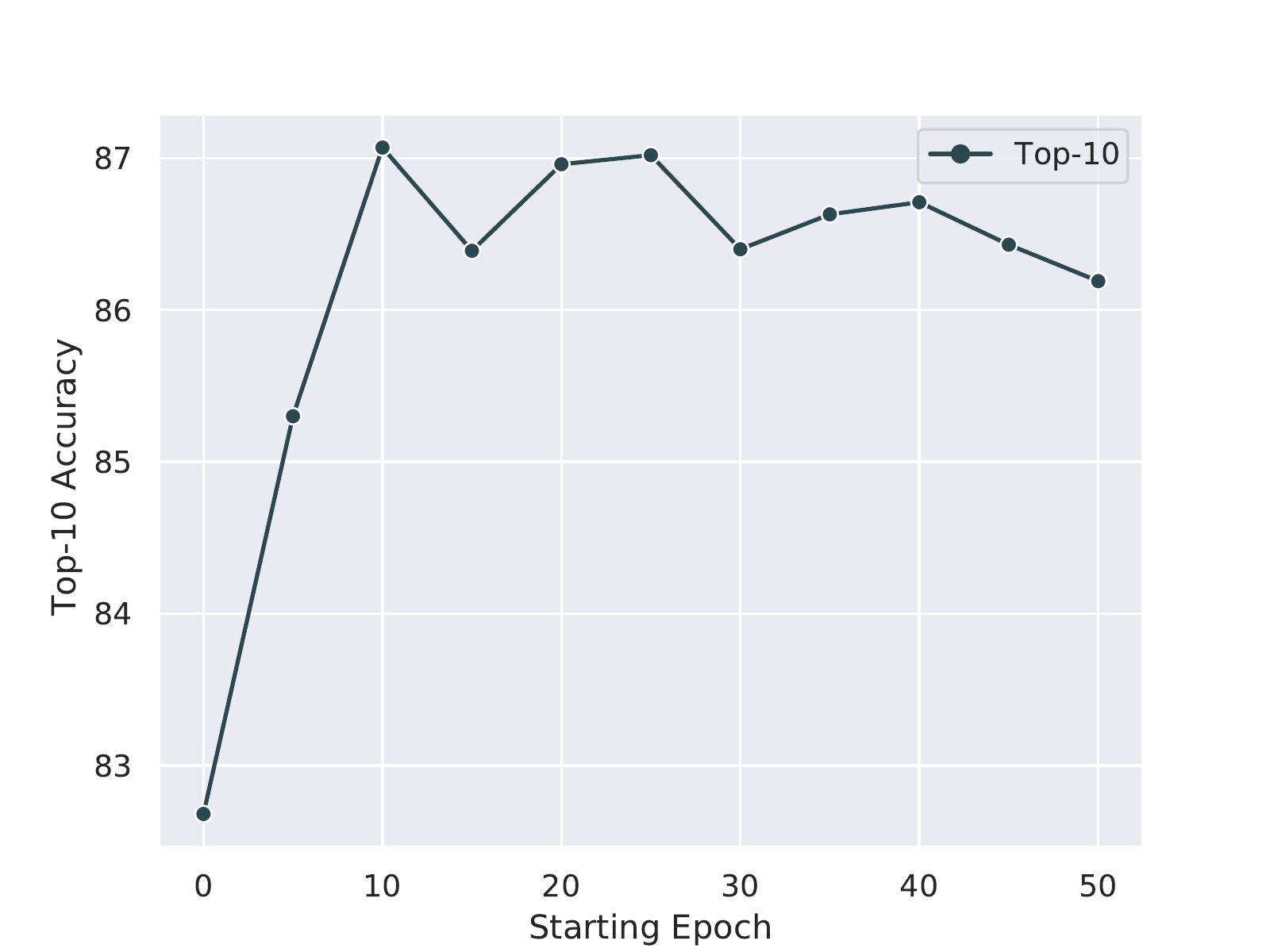}
		\end{minipage}%
	}%
	\centering
	\caption{Illustration of the impact of the starting epoch for the second training stage on CUHK-PEDES.}
	\label{fig:pt}
\end{figure*}

\subsubsection{\textbf{Impact of Distribution Shifting Mechanism}}
As shown in Tab.~\ref{tab:alba0}, the Distribution Shifting ($\mathcal{DS}$) mechanism is a key component in XProj, without which the performance drops by $1.71\%$, $1.04\%$, $0.87\%$ and $2.10\%$, $1.70\%$, $0.95\%$ on CUHK-PEDES and RSTPReid, respectively. The decrease in retrieval accuracy indicates the significance of the proposed $\mathcal{DS}$ mechanism.

\begin{figure}[!ht]
	\centering
	\includegraphics[width=\linewidth]{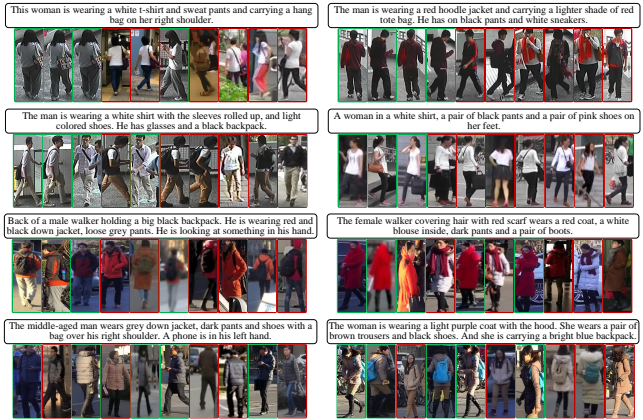}
	\caption{Illustration of top-10 text-based person retrieval results by LBUL. The matched pedestrian images are marked by green rectangles, while the mismatched person images are marked by red rectangles.}
	\label{fig:results-main}
\end{figure}

\subsubsection{\textbf{Impact of the Two-stage Training Strategy}}
As described in Sec.~\ref{sec:opti}, a two-stage strategy is proposed to train LBUL. To prove the validity of this optimization strategy, we conduct experiments which train LBUL with a one-stage strategy, namely, utilize $\mathcal{L}^{Stage2}$ since the beginning. As can be seen from Tab.~\ref{tab:alba0}, the retrieval performance decrease by $8.45\%$, $6.11\%$, $4.51\%$ and $8.50\%$, $6.70\%$, $5.55\%$ on CUHK-PEDES and RSTPReid, respectively, which demonstrate that after reliable uni-modal sub-manifolds are learned, a more consistent cross-modal common manifold can be constructed by LBUL. To further analysis the impact of the starting epoch for the second training stage, we conduct extensive experiments on CUHK-PEDES and the results are illustrated in Fig.~\ref{fig:pt}. It can be observed that initially the performance keeps increasing with the growth of the starting epoch for the second stage. Then after the value of the starting epoch passes 15, the performance gradually stabilizes. And the performance drop slightly after the value of the starting epoch gets too large. Note that when the starting epoch is 0, it is just equivalent to adopting a one-stage training strategy.

\subsection{Comparison with SOTA on Text-based Person Retrieval}

We compare the proposed LBUL with previous methods on CUHK-PEDES and RSTPReid. It can be observed from Tab.~\ref{tab:sota} and Tab.~\ref{tab:sota_retpreid} that without BERT, our proposed LBUL achieves $61.95\%$, $81.16\%$ and $87.19\%$ of rank-1, rank-5 and rank-10 accuracies respectively on CUHK-PEDES and $43.35\%$, $66.85\%$ and $76.50\%$ on RSTPReid. By encouraging the proposed model to look before it leaps with the proposed two-step common manifold mapping mechanism, LBUL outperforms existing methods and achieves the state-of-the-art performance on the text-based person retrieval task. For instance, TIMAM \cite{ARL} is one of the typical CDCP-based approaches. It aims to learn modality-invariant feature representations using adversarial and cross-modal matching objectives and utilizes a pretrained ResNet-101 as the visual backbone. With a ResNet-50 backbone, LBUL outperforms TIMAM by $7.44\%$, $3.60\%$ and $2.38\%$ on CUHK-PEDES of rank-$1$, rank-$5$ and rank-$10$ accuracies, respectively, which further proves the effectiveness of our proposed method. We display some examples of the top-$10$ text-based person retrieval results by LBUL in Fig.~\ref{fig:results-main}. The matched/mismatched pedestrian images are marked by green/red rectangles.

\section{Conclusion}
In this paper, we propose a novel algorithm termed LBUL to learn a Consistent Cross-modal Common Manifold (C$^{3}$M) for text-based person retrieval to overcome the CDCP dilemma. The core idea of our method, just as a Chinese saying goes, is `san si er hou xing', namely, to Look Before yoU Leap (LBUL). The common manifold mapping mechanism of LBUL includes two steps, namely, a looking step and a leaping step. Compared to CDCP-based common manifold mapping paradigms, LBUL considers distribution characteristics of both the visual and textual modalities before embedding data from one certain modality into C$^{3}$M to achieve a more solid cross-modal distribution consensus, and hence achieve a superior retrieval accuracy. We evaluate our proposed method on two text-based person retrieval datasets CUHK-PEDES and RSTPReid. Experimental results demonstrate that the proposed LBUL outperforms previous methods and achieves the state-of-the-art performance.

\begin{acks}
This work is partially supported by the National Natural Science Foundation of China (Grant No. 62101245, 61972016), China Postdoctoral Science Foundation (Grant No.2019M661999) , Natural Science Research of Jiangsu Higher Education Institutions of China (19KJB520009) and  Future Network Scientific Research Fund Project (Grant No. FNSRFP-2021-YB-21).
\end{acks}


\newpage
\bibliographystyle{ACM-Reference-Format}
\balance
\bibliography{jonniewayy}

%
%
%
%
%
%
%
%

\end{document}